\documentclass[10pt,twocolumn,letterpaper]{article}

\usepackage[pagenumbers]{cvpr}              

\usepackage{bm}
\usepackage{pifont} 
\usepackage{tabularx}
\usepackage{amsmath}
\usepackage{arydshln}
\newcommand{\cmark}{\textcolor{ForestGreen}{\ding{51}}}
\newcommand{\xmark}{\textcolor{red}{\ding{55}}}

\usepackage{xcolor}
\usepackage{soul}
\usepackage{todonotes}

\definecolor{light_gray}{rgb}{0.7, 0.7, .7}

\usepackage{booktabs}
\usepackage{setspace}
\usepackage[linesnumbered,ruled,vlined]{algorithm2e}
\usepackage{multirow}
\usepackage{graphicx}
\usepackage{array}
\usepackage{multirow}
\usepackage{graphicx}
\usepackage{subcaption}
\usepackage{array}
\usepackage{colortbl}
\usepackage{bbding}  
\usepackage{algorithmic}
\usepackage{longtable}
\usepackage{booktabs}

\definecolor{gray}{gray}{0.85}
\SetKwInput{KwInput}{Input}    
\SetKwInput{KwOutput}{Output}              %

\newcommand{\mypara}[1]{\noindent\textbf{{#1}}\xspace}

\newcommand{\name}{ZeroReg\xspace}

\newcommand{\red}[1]{\textcolor{red}{#1}}

\definecolor{cusyellow}{rgb}{1, 0.706, 0}
\definecolor{cusblue}{rgb}{0, 0.651, 0.929}

\definecolor{cvprblue}{rgb}{0.21,0.49,0.74}
\usepackage[pagebackref,breaklinks,colorlinks,allcolors=cvprblue]{hyperref}

\def\paperID{3357} 
\def\confName{CVPR}
\def\confYear{2025}

\title{ZeroReg: Zero-Shot Point Cloud Registration with Foundation Models}

\author{
  Weijie Wang\textsuperscript{1,2}\hphantom{,}
  Wenqi Ren\textsuperscript{1,3}\hphantom{,}
  Guofeng Mei\textsuperscript{2}\hphantom{,}
  Bin Ren\textsuperscript{1,4}\hphantom {,} \\
  Xiaoshui Huang\textsuperscript{5}
  Fabio Poiesi\textsuperscript{2}\hphantom{,}
  Nicu Sebe\textsuperscript{1}\hphantom{,}
  Bruno Lepri\textsuperscript{2},
  \\
  {\tt\small weijie.wang@unitn.it,
  gmei@fbk.eu, 
  }
  \\
  \textsuperscript{1}University of Trento, Italy 
  \quad 
  \textsuperscript{2}Fondazione Bruno Kessler, Italy \\
  \textsuperscript{3}East China University of Science and Technology, China \\
  \textsuperscript{4} Computer Vision Laboratory, ETH Zurich \quad
  \textsuperscript{5}Shanghai AI Laboratory, China \\
}

\begin{document}

\maketitle
\begin{abstract}
State-of-the-art 3D point cloud registration methods rely on labeled 3D datasets for training, which limits their practical applications in real-world scenarios and often hinders generalization to unseen scenes.
Leveraging the zero-shot capabilities of foundation models offers a promising solution to these challenges. 
In this paper, we introduce \textbf{\name}, a \textbf{zero}-shot \textbf{reg}istration approach that utilizes 2D foundation models to predict 3D correspondences.
Specifically, \name adopts an object-to-point matching strategy, starting with object localization and semantic feature extraction from multi-view images using foundation models.
In the object matching stage, semantic features help identify correspondences between objects across views. 
However, relying solely on semantic features can lead to ambiguity, especially in scenes with multiple instances of the same category. To address this, we construct scene graphs to capture spatial relationships among objects and apply a graph matching algorithm to these graphs to accurately identify matched objects.
Finally, computing fine-grained point-level correspondences within matched object regions using algorithms like SuperGlue and LoFTR achieves robust point cloud registration. Evaluations on benchmarks such as 3DMatch, 3DLoMatch, and ScanNet demonstrate \name's competitive performance, highlighting its potential to advance point-cloud registration by integrating semantic features from foundation models.

\end{abstract}    
\section{Introduction}
\label{sec:intro}

\begin{figure}[!t]
\centering
\resizebox{\linewidth}{!}{\includegraphics{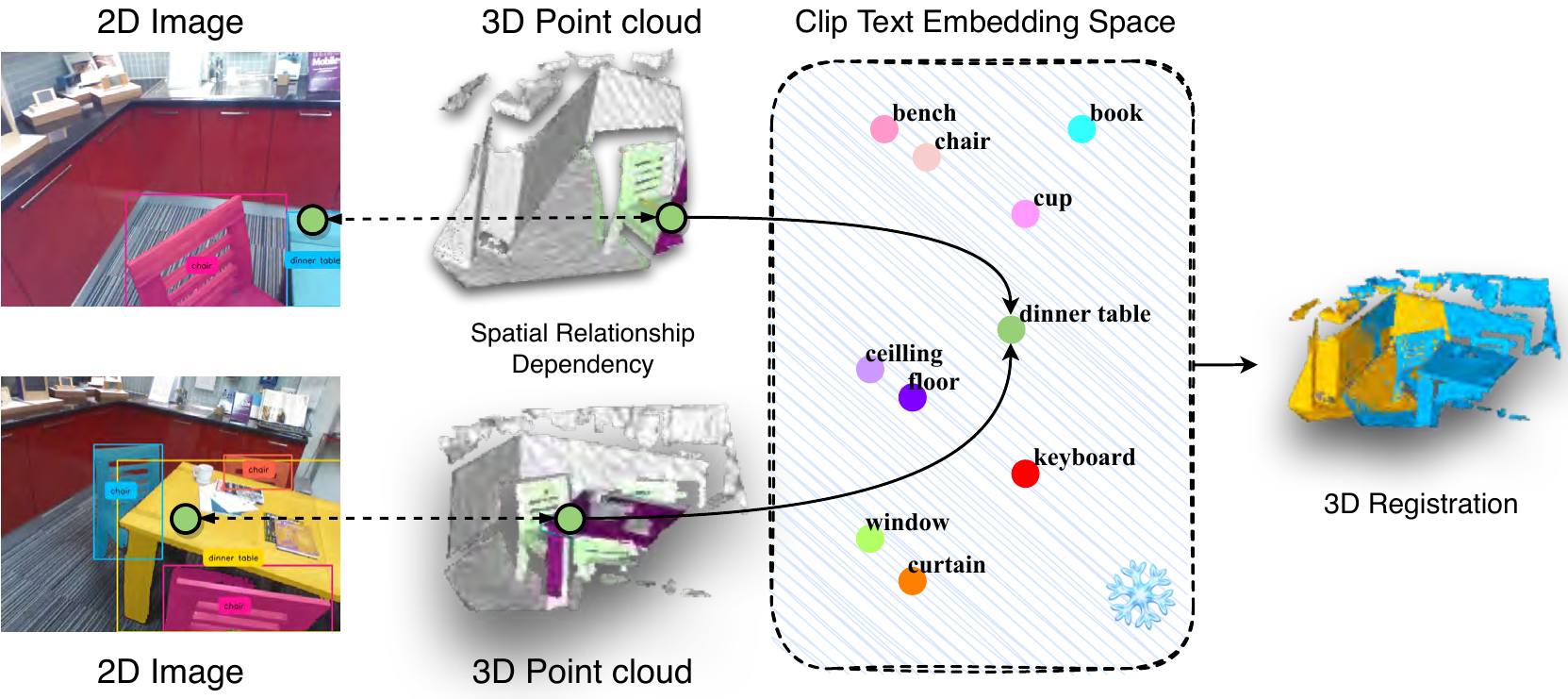}}
\caption{
\textbf{Key Motivation.} 
(1) We achieve zero-shot PCR by identifying matched objects based on semantic similarity using CLIP~\cite{radford2021learning}, which is pretrained on a large-scale image-text dataset.
(2) An object-centric scene graph is constructed to capture spatial relationships within the point clouds, resolving semantic ambiguities caused by multiple instances of the same category. Notably, our approach does not require additional 3D data training.
}
\label{sec:intro:fig1}
\vspace{-.4cm}
\end{figure}

Point cloud registration (PCR) aims to estimate the rigid transformation between a source and target point cloud~\cite{ao2020SpinNet}. 
While handcrafted methods~\cite{Besl1992, yang2020teaser, rusu2009fast} with manually engineered features remain effective in specific applications—particularly where data is limited or geometric constraints are clear, deep learning-based~\cite{huang2020feature, zhang2023pcrcg, banani2021unsupervisedrr, mei2023overlap} approaches are increasingly favored for their ability to learn complex and expressive feature representations. 
However, these learning-based methods require extensive training on 3D data, limiting their practical applications in real-world scenarios due to the high cost and time involved in acquiring annotated data. 

In contrast, humans inherently possess zero-shot capabilities, using contextual cues and prior knowledge to locate objects, infer spatial relations, and align scenes across different views. 
Equipping models with similar abilities could reduce dependence on labeled data and extensive training. 
One approach is to leverage large-scale pre-trained models like CLIP~\cite{radford2021learning}, which integrate contextual information and transfer knowledge across domains, enabling task-free adaptation to new situations. While CLIP has shown success in zero-shot visual recognition~\cite{li2023blip}, no 3D foundation model yet matches the zero-shot capabilities of its 2D counterparts, largely due to the limited availability of labeled 3D data.
Recent works~\cite{peng2023openscene, zhang2022pointclip, jatavallabhula2023conceptfusion, Zhu_2023_ICCV} attempted to explore transferring semantic features from foundation models, effectively advancing point cloud understanding in tasks like classification, segmentation, and detection. 

However, zero-shot PCR remains underexplored.
A key advantage is its ability to adapt to new scenes without additional 3D data training.
This motivated us to explore PCR in a zero-shot manner.
Moreover, since foundation models can distinguish different object categories, we are inspired to leverage these models to identify object-level correspondences.
However, achieving zero-shot PCR still entails two main challenges:
(\textbf{C1}) \textbf{Semantic Ambiguities}: In scenes with multiple instances of the same category, semantic ambiguities arise, e.g., when multiple chairs appear in the source, only one chair in the target—affected by viewpoint differences—corresponds correctly.
But which chair is the right match?
(\textbf{C2}) \textbf{Feature Similarity Within Same-Class Objects}:
The foundation models extract highly similar semantic features for objects within the same category, making it challenging to establish point-level correspondences, which is essential to PCR.

Based on these challenges, we introduce \name, a zero-shot registration method that leverages 2D foundation models to predict 3D correspondences. 
As illustrated in \cref{sec:intro:fig1}, \name facilitates PCR by utilizing the semantic capabilities of foundation models for object localization and scene graphs for accurate matching.
Specifically, we leverage foundation models, Florence-2~\cite{xiao2024florence} and SAMv2~\cite{ravi2024sam}, to detect and segment common objects within the source and target frames. 
CLIP~\cite{radford2021learning} is used to extract semantic features for matching detected objects across frames.
To address (\textbf{C1}), we construct a scene graph for each point cloud, where object centroids are represented as nodes, and edges capture their positional relationships. 
CLIP feature similarities are used as edge weights, providing a nuanced measure of similarity beyond simple binary values (0 or 1). 
This allows the graph to more accurately reflect relationships between objects.
We then predict object-level correspondences by solving a graph matching problem. 
Once object pairs are matched, we identify point-level correspondences within each matched object pair. 
To address \textbf{(C2)}, we utilize off-the-shelf feature-matching techniques, such as SuperGlue~\cite{sarlin2020superglue} and LoFTR~\cite{sun2021loftr}, for precise point-level matching. 
Finally, we employ RANSAC to calculate the transformation for registration.
\name is benchmarked against three datasets: 3DMatch, 3DLoMatch, and ScanNet, and it demonstrates competitive performance. 
Notably, \name outperforms hand-crafted methods and rivals several learning-based approaches, demonstrating its strong potential for PCR.

Compared to existing PCR methods, \name offers distinct advantages in data-scarce scenarios and when labeled data is costly, as shown in \cref{tab:intro:diff}. 
Its ease of use and scalability support broader adoption in 3D registration tasks. Additionally, leveraging foundation models for object localization and scene graphs is beneficial to PCR.

\begin{table}[!tb]
\centering
\resizebox{\linewidth}{!}{
\begin{tabular}{lccccc}
\hline
\multicolumn{1}{c}{\multirow{2}{*}{Method}} & No training & Semantic &  Robust to  &  Distinctive feature\\ 
\multicolumn{1}{c}{}  &  for 3D data  & awareness & noisy & representation capability   \\ \hline 
Handcrafted       & \cmark  & \xmark   & \xmark         & \xmark    \\
Learning-Based    & \xmark  & \xmark & \cmark    & \cmark  \\
\name       & \cmark   & \cmark   & \cmark    & \cmark         \\ \hline
\end{tabular}}
\vspace{-0.2cm}
\caption{
Comparison of existing PCR methods with \name. Handcrafted methods lack semantic awareness, robustness to noise, and distinctive feature representation. Learning-based methods, while capable of handling noise and providing distinctive feature representation, require 3D training data and still lack semantic awareness.  Contrastly, \name offers robustness, semantic awareness, and distinctive feature representation without requiring 3D data for training.
}
\vspace{-0.6cm}
\label{tab:intro:diff}
\end{table}
To summarize, our contributions are as follows:
\begin{itemize}
    \item We propose \name, the first method to leverage semantic features from foundation models in a zero-shot manner for PCR, eliminating the need for additional training on 3D data.
    \item We construct an object-matching scene graph based on KNN for point clouds to improve object-level reliability by integrating spatial relationships among objects, leveraging semantic features and positional context for more precise and reliable object matching, especially in scenes with multiple objects of the same class.
    \item We conduct extensive experiments on three benchmark datasets: 3DMatch, 3DLoMatch, and ScanNet, demonstrating competitive performance and highlighting the potential of leveraging semantic features from foundation models for PCR.
\end{itemize}
\section{Related Work}
\label{sec:related}
To achieve accurate PCR, numerous studies have focused on extracting discriminative features for correspondence prediction. 
Existing feature extraction methods for 3D PCR can be broadly categorized into \textbf{single-modal} and \textbf{multi-modal} approaches. 
We first discuss single-modal methods, followed by multi-modal ones. 
Since our approach leverages foundation models for correspondence search, we also review point cloud techniques using foundation models.
%
\paragraph{Single-Modal Methods} extract features directly from point cloud coordinates. Early methods relied on handcrafted features~\cite{drost2010model, rusu2009fast, johnson1999using, cirujeda2014mcov, aiger20084, salti2014shot}, which can be categorized into LRF-based and LRF-free approaches~\cite{deng2018ppfnet, rusu2008aligning, rusu2009fast}. LRF-based methods use Local Reference Frames but are sensitive to noise, while LRF-free methods, such as PPF, PFH, and FPFH, leverage geometric relationships to capture spatial configurations. Both approaches face challenges in complex, noisy environments.
Recent learning-based methods~\cite{Deng2018eccv, yang2018foldingnet, ao2020SpinNet, deng2018ppfnet, huang2021predator, yu2021cofinet, qin2023geotransformer, bai2020d3feat, gojcic2019perfect} significantly improve upon handcrafted features by learning more robust representations. For instance, PPF-FoldNet~\cite{Deng2018eccv} encodes PPF patches and reconstructs them using FoldingNet~\cite{yang2018foldingnet}. D3Feat~\cite{bai2020d3feat} and Predator~\cite{huang2021predator} focus on detecting reliable keypoints for correspondence, while CoFiNet~\cite{yu2021cofinet} avoids keypoint detection, extracting coarse-to-fine correspondences instead. PerfectMatch~\cite{gojcic2019perfect} standardizes patches with LRF to enhance rotation robustness, and SpinNet~\cite{ao2020SpinNet} aligns local patches before feature learning. YOHO~\cite{wang2022you} trains features using a dodecahedral set, but its rotational properties are fragile due to the limitations of finite rotation representation. GeoTransformer~\cite{qin2023geotransformer} learns geometric features for robust superpoint matching, making it effective in low-overlap and rigid transformation scenarios.

\paragraph{Multi-Modal Methods} leverage RGB-D images or combine geometric and visual features to enhance PCR performance. 
Compared to single-modal methods, integrating visual features from RGB images with geometric features provides greater discriminative power~\cite{Kang_2024, Hatem_2023_ICCV, Liu_2023_ICCV, Liu1_2023_ICCV}. 
Early efforts in multimodal feature extraction, such as~\cite{bay2006surf, lowe2004distinctive, rublee2011orb}, relied on patch-based 2D visual features combined with geometric features from local 3D relationships. 
Over time, these approaches evolved to use convolutional neural networks for extracting more robust visual and geometric features~\cite{moravec1981rover}.
Recent methods~\cite{banani2021unsupervisedrr, wang2022improving} often utilize geometric transformations to learn visual features or establish correspondences with large-scale image datasets. For example, UnsupervisedR\&R~\cite{banani2021unsupervisedrr} uses pose transformations within RGB-D video data, extending multimodal efforts through supervised pose augmentation. Building on this, LLT~\cite{wang2022improving} introduces a multi-scale local linear transformation technique to combine visual and geometric features from RGB and depth images, addressing visual disparities caused by geometric variations.
However, existing learning-based techniques require extensive training on 3D data, which limits their practicality. 
In contrast, our work explores leveraging semantic features from foundation models to perform PCR without additional 3D data training.

\paragraph{Vision Understanding via Foundation Models.}
Recently, numerous foundation models~\cite{radford2021learning, jia2021scaling, xiao2024florence, ravi2024sam, kirillov2023segment} have been proposed, demonstrating impressive zero-shot transfer capabilities for tasks such as classification, segmentation~\cite{ravi2024sam, kirillov2023segment}, detection~\cite{xiao2024florence}, and vision-language understanding~\cite{openclip, radford2021learning}. 
These capabilities are largely attributed to pretraining on large-scale datasets, enabling these models to capture general semantic features and transfer effectively to various downstream tasks.
Inspired by these advancements, researchers explore similar zero-shot capabilities to 3D point cloud understanding~\cite{ma2024shapesplat, ren2024bringing, peng2023openscene, zhang2022pointclip}. 
PointCLIP~\cite{zhang2022pointclip} pioneers the use of CLIP for 3D tasks, leveraging point cloud depth maps with CLIP’s visual encoder for zero-shot classification.
Openscene~\cite{peng2023openscene} proposes a zero-shot 3D scene understanding method that co-embeds 3D points, text, and image pixels in the CLIP feature space, enabling task-agnostic training and open-vocabulary querying. 
These methods extend CLIP’s zero-shot capabilities to point cloud understanding by embedding 3D points, text, and images into its feature space, enabling task-agnostic 3D interpretation. 
Our approach leverages semantic features from foundation models for PCR in a zero-shot manner, which in turn reduces the reliance on 3D data while achieving robust registration performance.
\section{Method}
\label{method}
\begin{figure*}[!t]
\centering  
\includegraphics[width=1\textwidth]{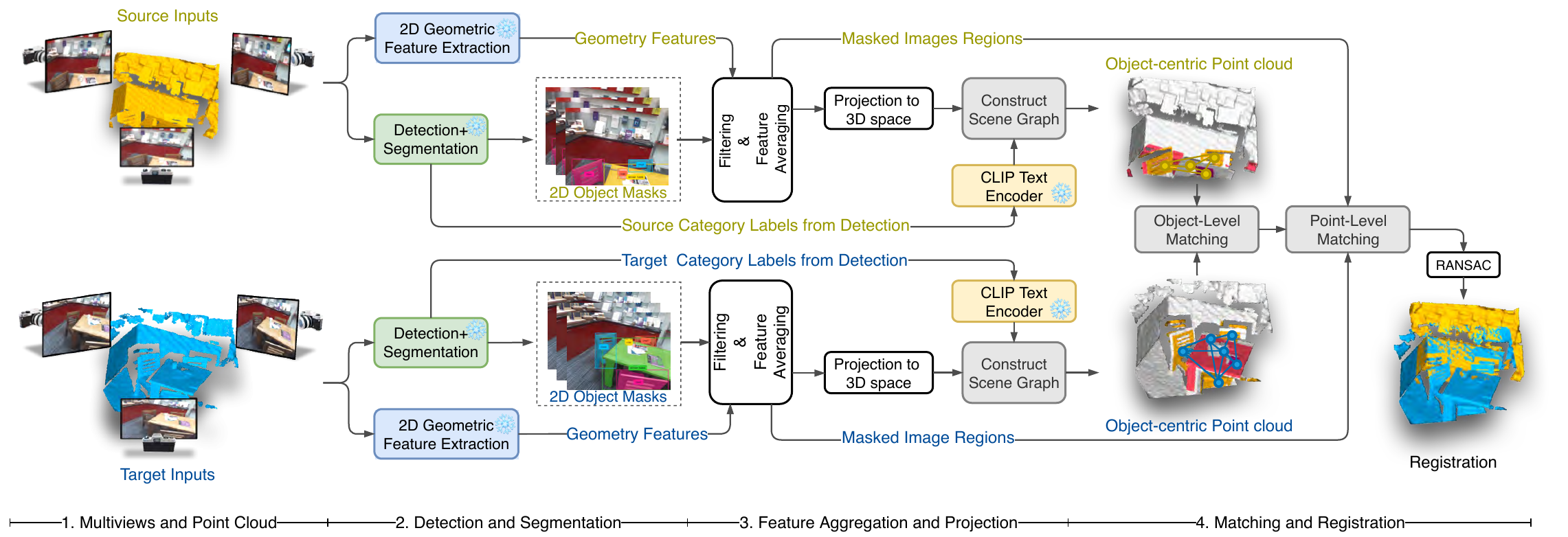}
\caption{The \name framework begins with segmentation and detection of source and target multiviews using foundational visual models, along with geometric feature extraction through feature matching algorithms. 
The generated multi-view object masks are then filtered and averaged, with semantic features extracted using the CLIP text encoder in parallel with geometric feature processing, before being projected onto the point cloud. 
A scene graph is constructed to facilitate object-level and point-level matching, establishing correspondences between the source and target. Based on these correspondences, the transformation is calculated, and RANSAC is applied for optimization, achieving precise registration.}

\vspace{-3mm}

\label{fig:frame}
\end{figure*}
This paper proposes a method that leverages foundation models to predict 3D correspondences for PCR, without the need for additional 3D data training.
Our approach is organized as follows: \cref{method:def} introduces the problem statement; \cref{method:clip_sem} explains the processes of object localization and feature extraction; \cref{method:object_match} describes object-level matching; and \cref{method:point_match} details the prediction of point-level correspondence for registration.

\subsection{Problem Statement} 
\label{method:def}
\cref{fig:frame} shows an overview of our \name. Given two partially overlapping point clouds, source \( \bm{\mathcal{P}} = \{\bm{p}_i \in \mathbb{R}^{3} \mid i = 1, 2, \dots, n\} \) and target \( \bm{\mathcal{Q}} = \{\bm{q}_i \in \mathbb{R}^{3} \mid i = 1, 2, \dots, m\} \), along with their corresponding multi-view images \( \mathcal{I} = \{\mathcal{I}_{\mathcal{P}}^i, \mathcal{I}_{\mathcal{Q}}^i \mid i = 1, 2, \dots, k\} \), the goal of registration is to estimate the rigid transformation \( \mathcal{T} \in SE(3) \).
\name leverages foundation models for detection~\cite{xiao2024florence, wang2024yolov10} and segmentation~\cite{ravi2024sam} to locate objects in the scene, generate category labels and masks from multi-view images. 
We use CLIP’s text encoder~\cite{radford2021learning} to embed each object's category label as a semantic feature. 
For geometric features extraction within the masked regions, we apply feature matching methods~\cite{detone2018superpoint, sarlin2020superglue, sun2021loftr}, such as SuperGlue and LoFTR, which is method-agnostic.
The semantic and geometric features are projected onto the point clouds. 
Using object centroids and semantic features, we then construct scene graphs for \( \bm{\mathcal{P}} \) and \( \bm{\mathcal{Q}} \), which serve as inputs to the object-matching block to establish object-level correspondences.
For each matched object pair, geometric features help identify point-level correspondences, which are used to conduct the registration via RANSAC~\cite{fischler1981random}.


\subsection{Object Localization and Feature Extraction}
\label{method:clip_sem}
The initial stage of \name involves localizing object positions, extracting semantic and geometric features, and back-projecting these features into the scene’s 3D space.

\mypara{Object Detection and Segmentation.} 
We use Florence-2~\cite{xiao2024florence} for detection and SAMv2~\cite{ravi2024sam} for segmentation to semantically localize the position of objects.
For the \( i \)-th image pair \( (\mathcal{\bm{I}}_{\mathcal{P}}^i, \mathcal{\bm{I}}_{\mathcal{Q}}^i) \), 
we first detect and localize objects to generate \textsc{Bounding Boxes} and \textsc{Category Labels}.
These bounding boxes serve as prompts for SAMv2~\cite{ravi2024sam} to produce \textsc{Object Masks}.
Now, we have masks as well as category labels, denoted \( \bm{C}_{\mathcal{P}}^{i,n} \) for $\mathcal{P}$ and \( \bm{C}_{\mathcal{Q}}^{i,m} \) for $\mathcal{Q}$, where \( n \) and \( m \) denote the number of detected objects.

\mypara{Semantic Feature Extraction.}
Once the objects in the scene are localized, we use semantic information to facilitate object matching. 
Our goal is to identify object-level correspondences between \( \bm{\mathcal{P}} \) and \( \bm{\mathcal{Q}} \) by computing the similarity using category labels, 
thereby establishing accurate object matches.
Specifically, given the \( i \)-th pair of images with category labels \( (\bm{C}_{\mathcal{P}}^{i,n}, \bm{C}_{\mathcal{Q}}^{i,m}) \), we independently extract semantic features for each category label 
using a pre-trained CLIP text encoder~\cite{radford2021learning}, 
resulting in semantic feature sets 
\( (\hat{\bm{S}}_{\mathcal{P}}^i, \hat{\bm{S}}_{\mathcal{Q}}^i) \).
For multiple views, we repeat the above operation. Importantly, we retain only masks that are visible in at least two views, ignoring objects detected in a single view. 
This would improve the robustness of the feature representation by focusing on regions that are consistently detected across multiple views.
For regions that overlap across multiple views, we calculate the semantic features of the overlapping regions and take their average to produce a unified representation, which ensures consistency and reduces redundancy. 
Other multi-view feature fusion methods can also be applied here; however, our focus is on finding matching objects, so we do not discuss feature fusion in detail in this context.
Specifically, let \( \hat{\bm{f}}_{i} \) represent feature vectors from overlapping regions across multiple views \( i = 1, \dots, k \), either from the source or target. The averaged feature vector across views is given by \( \tilde{\bm{f}}_{\text{avg}} = \frac{1}{k} \sum_{i=1}^{k} \hat{\bm{f}}_{i} \).
This process ultimately produces the semantic feature sets \( \tilde{\bm{S}}_{\mathcal{P}} \) for the source and \( \tilde{\bm{S}}_{\mathcal{Q}} \) for the target across different views, which can be used for further alignment or analysis in subsequent steps.

\mypara{Geometric Feature Extraction.}
For the \( i \)-th pair of images \( (\mathcal{\bm{I}}_{\mathcal{P}}^i, \mathcal{\bm{I}}_{\mathcal{Q}}^i) \),
we utilize standard feature matching methods, such as SuperGlue~\cite{sarlin2020superglue} and LoFTR~\cite{sun2021loftr}, to extract geometric features, consisting of position and context-dependent information, from the masked image pairs \( (\bm{\mathcal{I}}_{\mathcal{P}}^i \odot \textsc{Mask}_{P}, \bm{\mathcal{I}}_{\mathcal{Q}}^i \odot \textsc{Mask}_{Q}) \).
The process focuses on the semantic regions, ignoring the irrelevant ones to extract reliable geometric features for the following processing.
For multiple views, we independently extract geometric features \( {\bm{G}}_{\mathcal{P}}^j \) and \( {\bm{G}}_{\mathcal{Q}}^k \) for each pair and average them in overlapping regions to create a unified feature, as discussed before.
Finally, we obtain the aggregated geometric features \( \tilde{\bm{G}}_{\mathcal{P}} \) and \( \tilde{\bm{G}}_{\mathcal{Q}} \) across different views.

\mypara{Object and Feature Projection.}
After aggregating the semantic feature sets \( (\tilde{\bm{S}}_{\mathcal{P}}, \tilde{\bm{S}}_{\mathcal{Q}}) \) and the geometric feature sets \( (\tilde{\bm{G}}_{\mathcal{P}}, \tilde{\bm{G}}_{\mathcal{Q}}) \), 
we project them onto 3D space using the camera parameters and depth images to obtain the final semantic feature sets \( \bm{S}_{\mathcal{P}} \) and \( \bm{S}_{\mathcal{Q}} \), as well as the geometric feature sets \( \bm{G}_{\mathcal{P}} \) and \( \bm{G}_{\mathcal{Q}} \) for \( \mathcal{P} \) and \( \mathcal{Q} \), respectively.
More details of the back-projection process are provided in the Appendix. 
Thus, the source $\mathcal{P}$ and target $\mathcal{Q}$, segmented by different masks and containing geometric features and labels, can be represented as follows:
\begin{equation*}
\begin{aligned}
    \bm{\mathcal{P}} &{=} \bigcup^{m}_{j=1}\bm{\mathcal{P}}^j,
    \bm{\mathcal{P}}^j {=} \left\{ \left( \bm{p}_i, c^j, \bm{s}^j \right) \mid i \in \bm{m}^j \right\}, j = 1, \dots, m \\
    \bm{\mathcal{Q}} &{=} \bigcup^{n}_{k=1}\bm{\mathcal{Q}}^k,
    \bm{\mathcal{Q}}^k {=} \left\{ \left( \bm{q}_i, c^k, \bm{s}^k \right) \mid i \in \bm{m}^k \right\}, k = 1, \dots, n
\end{aligned}
\end{equation*}
where:
\begin{itemize}
    \item \(\bm{\mathcal{P}}^j\), \(\bm{\mathcal{Q}}^k\): subsets of points defined by the masks \(\bm{m}^j\) and \(\bm{m}^k\), each associated with category labels \(c_j^p\) and \(c_k^q\).
    \item \(\bm{p}_i, \bm{q}_i \in \mathbb{R}^3\): 3D coordinate points in the sets \(\bm{\mathcal{P}}\) and \(\bm{\mathcal{Q}}\).
    \item \(\bm{s}_j, \bm{s}_k \in \mathbb{R}^d\): unified semantic feature vectors representing each subset \(\bm{\mathcal{P}}^j\) and \(\bm{\mathcal{Q}}^k\).
    \item \(m\) and \(n\): total number of masks (or subsets) in \(\bm{\mathcal{P}}\) and \(\bm{\mathcal{Q}}\).
\end{itemize}
Also, for the geometric features, we define \(\{\bm{g}^{p}_{l}\}_{l=1}^{L_P}\) and \(\{\bm{g}^{q}_{l}\}_{l=1}^{L_Q}\) as the sets of global geometric features, where \( L_P \) and \( L_Q \) denote the number of geometric feature points in the sets \(\bm{\mathcal{P}}\) and \(\bm{\mathcal{Q}}\), respectively.

\subsection{Graph-Based Object-Level Matching}
\label{method:object_match}
To address semantic ambiguity, we construct scene graphs for objects within point clouds to capture spatial and relational context. Leveraging these positional relationships, we match multiple objects between the source and target point clouds by transforming object matching into a search for node-to-node correspondences between the two graphs.

Specifically, we first construct two scene graphs, $\mathcal{G}_\mathcal{P}$ and $\mathcal{G}_\mathcal{Q}$, from the masked point clouds obtained in \cref{method:clip_sem}. Let $\mathcal{G}_\mathcal{P} = (\mathcal{V}^p, \mathcal{W}^p)$, where $|\mathcal{V}^p| = n$ for $\bm{\mathcal{P}}$, and $\mathcal{G}_\mathcal{Q} = (\mathcal{V}^q, \mathcal{W}^q)$, where $|\mathcal{V}^q| = m$ for $\bm{\mathcal{Q}}$, where $\mathcal{V}$ represents the set of nodes and $\mathcal{W}$ is the affinity matrix.
For each node \( \mathcal{V}^{p}_j \) in the graph \( \mathcal{G}_\mathcal{P} \), we compute the centroid of its corresponding point set in $\bm{\mathcal{P}}$, denoted as \( \bm{v}^{p}_j = \frac{1}{n} \sum_{i=1}^{n} \bm{p}_{i} \). 
We apply the same operation for $\bm{\mathcal{Q}}$, obtaining \( \bm{v}_k^{q} = \frac{1}{m} \sum_{i=1}^{m} \bm{q}_{i} \).
The affinity matrices \( \mathcal{W}^p \in \mathbb{R}^{n \times n} \) and \( \mathcal{W}^q \in \mathbb{R}^{m \times m} \) are constructed using $ k $-nearest neighbors (kNN), capturing local spatial relationships to maintain neighborhood consistency. The elements of matrices are calculated by:
\begin{equation}
    \mathcal{W}^x_{jk} =
    \begin{cases}
        1 + \frac{{\bm{s}^x_j}^\top \bm{s}^x_k}{\|\bm{s}^x_j\| \|\bm{s}^x_k\|}, & \text{if } v^x_k \in \text{kNN}(v^x_j), \\
        0, & \text{otherwise}, 
    \end{cases}
    \label{method:eq:2}
\end{equation}
we replace binary edge weights (0 or 1) with semantic feature similarities, enabling a more nuanced differentiation between neighboring nodes.
Furthermore, we consider the semantic relationship between the $j$-th node $v^p_j$ in $\bm{\mathcal{P}}$ and the $k$-th node $v^q_k$ in $\bm{\mathcal{Q}}$, calculated as $\mathbf{C} \in \mathbb{R}^{n \times m} = \{1 + \frac{{\bm{s}^s_j}^\top \bm{s}^t_k}{\|\bm{s}^s_j\| \|\bm{s}^t_k\|} \mid 1 \leq j \leq n, 1 \leq k \leq m \}$. This matrix $\mathbf{C}$ measures node similarities between the two graphs.

Let $\mathbf{X}$ denote the assignment matrix representing the matching between graphs $\mathcal{G}_\mathcal{P}$ and $\mathcal{G}_\mathcal{Q}$, where $X_{jk} = 1$ indicates that the $j$-th node $v^p_j$ in $\bm{\mathcal{P}}$ matches the $k$-th node $v^q_k$ in $\bm{\mathcal{Q}}$, and $X_{jk} = 0$ otherwise. 
Since nodes may represent similar objects, they can form complex many-to-many relationships across graphs, which is crucial for accurately matching multiple similar objects in both source and target point clouds.
The graph matching problem is formulated as:
\begin{equation}\label{eq:graph_match}
    \begin{aligned}
        & \min_{\mathbf{X}} \| \mathcal{W}^p - \mathbf{X} \mathcal{W}^q \mathbf{X}^\top \|_F^2 - \operatorname{tr}(\mathbf{C} \mathbf{X}), \\
        & \mathbf{X} \in \{0, 1\}^{n \times m}, \quad \mathbf{X} \mathbf{1}_n = \mathbf{1}_m, \quad \mathbf{X}^\top \mathbf{1}_n \leq \mathbf{1}_n,
    \end{aligned}
\end{equation}
The first term minimizes structural differences between the graphs, while the second term maximizes node similarity based on semantic features. Generally, \cref{eq:graph_match} represents a quadratic assignment problem~\cite{loiola2007survey}.

By solving \cref{eq:graph_match}, we obtain object-level correspondences $ \mathcal{C}_{\text{object}} = \{(O^j, O^k)\}_{i=1}^L $, where $ L $ denotes the total number of matched pairs. We then filter out pairs that do not belong to the same category using category labels from the detection phase (\cref{method:clip_sem}), ensuring consistency and improving accuracy when matching multiple similar objects.

\subsection{Semantic-Guided Point-Level Matching}
\label{method:point_match}
After obtaining the object-level correspondences \( \mathcal{C}_{\text{object}} \), we proceed to search for the final point-level correspondences \( \mathcal{C}_{\text{point}} \), utilizing geometric features (\(\mathbf{G}_\mathcal{P}\), \(\mathbf{G}_\mathcal{Q}\)) to complete the process. 
We focus on the regions specified by matched object pairs to obtain point-level correspondences.
The similarity matrix is defined as follows:
\begin{equation}
\mathbf{S}_{\text{sim}} = \mathbf{G}_{\mathcal{P}} \cdot (\mathbf{G}_{\mathcal{Q}})^T,
\end{equation}
where the similarity matrix \( \mathbf{S}_{\text{sim}} \) represents the preliminary point-level matching, and $\cdot$ denotes the inner product operation. 
\( \mathbf{S}_{\text{sim}} \) captures the geometric relationship between the source and target point clouds.
Semantic guidance significantly enhances point-level matching by focusing on well-aligned regions. Building on this, the integration of semantic and geometric similarities ensures consistent and meaningful matches, even when geometric information alone is insufficient.
To further improve matching flexibility, we augment \( \mathbf{S}_{\text{sim}} \) by adding a new row and a new column as described in~\cite{7406423} and compute it as outlined in \cref{eq:slack}:
\begin{equation}
\label{eq:slack}
\mathbf{S^{\prime}} =\begin{bmatrix}
\mathbf{S}_{\text{sim}} & \mathbf{z} \\
\mathbf{z}^T & z
\end{bmatrix}, \quad \mathbf{S^{\prime}} \in \mathbb{R}^{(n^{\prime}+1) \times (m^{\prime}+1)},
\end{equation}
where \( z = 1 \) and \(\bm{z}\) is an auxiliary vector with all elements set to 1. Note that \( n^{\prime} \) and \( m^{\prime} \) are not fixed constants but vary dynamically based on the size of the matched regions, reflecting the focus on localized areas within matched objects.
This augmentation effectively introduces a ``slack'' element that allows the similarity matrix to accommodate unmatched points, thereby improving flexibility and aiding in a more robust optimization process.
Next, we apply the Sinkhorn algorithm~\cite{cuturi2013sinkhorn} to \( \mathbf{S^{\prime}} \) to solve the optimal transport problem, enhancing point-level matching. 
After that, we drop the last row and column to yield the final correspondences \( \mathcal{C}_{\text{point}} \). 
We then use the RANSAC algorithm~\cite{rusu2009fast} to conduct registration.
The \emph{pseudo-code} of the entire framework is provided in the Appendix.

\section{Experiments}
\label{experiment}
Our experiments are organized as follows:
In ~\cref{exp:exp_setup} we describe the experimental setup, including datasets, implementation details, and evaluation metrics.
We then evaluate \name with current SOTA learning-based, hand-crafted, and zero-shot methods on 3DMatch~\cite{zeng20173dmatch} and 3DLoMatch~\cite{huang2021predator} in ~\cref{exp:comparisons_3dm_3dlm} and ScanNet~\cite{dai2017scannet} in ~\cref{exp:comparisons_scan}, comparing it with current SOTA learning-based.
Finally, we conduct ablation studies and analysis in \cref{exp:ablation}.

\subsection{Experiments Setup}
\label{exp:exp_setup}
\mypara{Dataset.}
(1) 3DMatch~\cite{huang2021predator} and 3DLoMatch datasets~\cite{zeng20173dmatch}: An indoor dataset with 62 scenes (46/8/8 for training/validation/testing). We focus on the test set, which includes 1623 point cloud fragments (3DMatch) and 1781 fragments (3DLoMatch), along with transformation matrices preprocessed by ~\cite{huang2021predator}. Their overlap regions exceed 30\% and range from 10\% to 30\%, respectively.
(2) ScanNet-v1~\cite{dai2017scannet}: This dataset includes  1045/156/312 scenes for training, validation, and testing. We generate 26K view pairs in the test set by selecting image pairs with a 20-frame interval.

\mypara{Implementation.}
\name\ is implemented in PyTorch~\cite{paszke2019pytorch} and Open3D~\cite{zhou2018open3d}. We use Open3D’s RANSAC for transformation estimation between point clouds, and conduct all experiments on a Tesla V100 GPU. RGB images are sized at 640×480, while input images for the CLIP ViT-B/32 model~\cite{radford2021learning} are resized to 224×224 with a patch size of 32×32. Detection and segmentation follow official setups for Florence-2-base~\cite{xiao2024florence} and SAM2\_base\_plus~\cite{ravi2024sam}. 
Feature matching algorithms and the Sinkhorn algorithm~\cite{cuturi2013sinkhorn} (20 iterations) also adhere to official configurations, with superpoint threshold $\gamma$ = 0.05. 
For the nearest neighbors, we select 3 as the value of $k$.
Additional implementation details are in the Appendix.

\mypara{Evaluation Metrics.}
We follow~\cite{yu2021cofinet} and use the root mean square error (RMSE) metric to compute the recall ratio (RR) and inlier ratio (IR) for evaluating registration performance on the 3DMatch and 3DLoMatch datasets, adopting RMSE $<$ 20cm as the success threshold. 
Additionally, we use the rotation error (RE) and translation error (TE) metrics from~\cite{banani2021unsupervisedrr} to evaluate registration performance on ScanNet, reporting translation error in centimeters and rotation error in degrees. 
More details on evaluation metrics are provided in the Appendix.


\subsection{Comparisons on 3DMatch \& 3DLoMatch}
\label{exp:comparisons_3dm_3dlm}
Our method \name focuses on the prediction of correspondence, thus our comparisons are limited to methods specifically related to the prediction of correspondence.
\begin{itemize}
    \item \textbf{Supervised}: We compare our approach with several supervised learning-based methods trained on 3D data, including PPFNet~\cite{deng2018ppfnet}, PerfectMatch~\cite{gojcic2019perfect}, FCGF~\cite{choy2019fully}, D3Feat~\cite{bai2020d3feat}, SpinNet~\cite{ao2020SpinNet}, YOHO~\cite{wang2022you}, RIGA~\cite{yu2022riga}, Predator~\cite{huang2021predator}, CoFiNet~\cite{yu2021cofinet}, GeoTransformer~\cite{qin2023geotransformer}, and RoITr~\cite{yu2023rotationinvariant}.

    \item \textbf{Unsupervised}: FoldingNet~\cite{yang2018foldingnet}, PPF-FoldNet~\cite{Deng2018eccv}, CapsuleNet~\cite{Zhao2019}, and Equivariant3D~\cite{Spezialetti2019}.

    \item \textbf{Handcrafted} and \textbf{Zero-shot}: FPFH~\cite{rusu2009fast}, USC~\cite{Tombari2010}, and FreeReg~\cite{wang2023freereg}, respectively.
\end{itemize}
Our method relies on point-level correspondences from feature matching across masked regions, yielding approximately 1000 points, with ``sample=1000’’ chosen for comparison as shown in ~\cref{exp:tab:3d_3dl}.
\begin{table}[t]
\renewcommand{\arraystretch}{1.0} 
\centering
\tabcolsep 3pt 
\caption{Registration Recall (RR) and Inlier Ratio (IR) on 3DMatch and 3DLoMatch. 
Our approach is compared against three different types of state-of-the-art (SOTA) 3D descriptors for point cloud registration. 
The \textbf{highest} performance is shown in bold.
For supervised ones, we choose the number of sampling points is \textbf{``sample=1000''} for comparison.
}
\resizebox{\columnwidth}{!}{%
\label{exp:tab:3d_3dl}
\begin{tabular}{lcccccc}
    \toprule[1pt]
    \multirow{2}{*}{\textbf{Method}} & \multirow{2}{*}{\shortstack{\textbf{Train on}\\\textbf{3DMatch}}} & \multicolumn{2}{c}{\textbf{3DMatch}} & \multicolumn{2}{c}{\textbf{3DLoMatch}} \\
    \cmidrule(lr){3-4} \cmidrule(lr){5-6} 
    & & \textbf{RR (\%)} & \textbf{IR (\%)} & \textbf{RR (\%)} & \textbf{IR (\%)} \\ 
    \midrule
    \multicolumn{6}{l}{\textbf{\emph{Hand-crafted methods}}} \\
    \cdashline{1-6}[2pt/1pt] 
    FPFH~\cite{rusu2009fast} & $\checkmark$ & \cellcolor{gray!50}40.3 & \cellcolor{gray!50}34.1 & \cellcolor{gray!50}- & \cellcolor{gray!50}- \\    
    USC~\cite{Tombari2010} & $\checkmark$ & \cellcolor{gray!50}43.2 & \cellcolor{gray!50}- & \cellcolor{gray!50}- & \cellcolor{gray!50}- \\
    \midrule
    \multicolumn{6}{l}{\textbf{\emph{Supervised learning-based methods}}} \\
    \cdashline{1-6}[2pt/1pt] 
    PPFNet~\cite{deng2018ppfnet} & $\checkmark$ & \cellcolor{gray!50}71.0 & \cellcolor{gray!50}- & \cellcolor{gray!50}- & \cellcolor{gray!50}- \\
    PerfectMatch~\cite{gojcic2019perfect} & $\checkmark$ & \cellcolor{gray!50}71.4 & \cellcolor{gray!50}- & \cellcolor{gray!50}23.3 & \cellcolor{gray!50}- \\
    FCGF~\cite{choy2019fully} & $\checkmark$ & \cellcolor{gray!50}83.3 & \cellcolor{gray!50}48.7 & \cellcolor{gray!50}38.2 & \cellcolor{gray!50}17.2 \\
    D3Feat~\cite{bai2020d3feat} & $\checkmark$ & \cellcolor{gray!50}83.4 & \cellcolor{gray!50}40.4 & \cellcolor{gray!50}46.9 & \cellcolor{gray!50}14.0 \\
    SpinNet~\cite{ao2020SpinNet} & $\checkmark$ & \cellcolor{gray!50}85.5 & \cellcolor{gray!50}40.8 & \cellcolor{gray!50}48.3 & \cellcolor{gray!50}20.6 \\
    YOHO~\cite{wang2022you} & $\checkmark$ & \cellcolor{gray!50}89.1 & \cellcolor{gray!50}55.7 & \cellcolor{gray!50}63.2 & \cellcolor{gray!50}22.6 \\
    RIGA~\cite{yu2022riga} & $\checkmark$ & \cellcolor{gray!50}89.1 & \cellcolor{gray!50}70.6 & \cellcolor{gray!50}\textbf{64.5} & \cellcolor{gray!50}34.3 \\
    Predator~\cite{huang2021predator} & $\checkmark$ & \cellcolor{gray!50}90.6 & \cellcolor{gray!50}57.1 & \cellcolor{gray!50}58.1 & \cellcolor{gray!50}28.3 \\
    CoFiNet~\cite{yu2021cofinet} & $\checkmark$ & \cellcolor{gray!50}88.4 & \cellcolor{gray!50}51.9 & \cellcolor{gray!50}60.7 & \cellcolor{gray!50}26.7 \\
    GeoTransformer~\cite{qin2023geotransformer} & $\checkmark$ & \cellcolor{gray!50}\textbf{91.8} & \cellcolor{gray!50}76.0 & \cellcolor{gray!50}59.9 & \cellcolor{gray!50}46.2 \\
    RoITr~\cite{yu2023rotationinvariant} & $\checkmark$ & \cellcolor{gray!50}\textbf{91.8} & \cellcolor{gray!50}\textbf{83.0} & \cellcolor{gray!50}\textbf{74.8} & \cellcolor{gray!50}\textbf{55.1} \\
    \midrule[0.8pt]
    \multicolumn{6}{l}{\textbf{\emph{Unsupervised learning-based methods}}} \\
    \cdashline{1-6}[2pt/1pt] 
    FoldingNet~\cite{yang2018foldingnet} & $\checkmark$ & \cellcolor{gray!50}68.3 & \cellcolor{gray!50}- & \cellcolor{gray!50}- & \cellcolor{gray!50}- \\
    PPF-FoldNet~\cite{Deng2018eccv} & $\checkmark$ & \cellcolor{gray!50}70.5 & \cellcolor{gray!50}- & \cellcolor{gray!50}- & \cellcolor{gray!50}- \\
    CapsuleNet~\cite{Zhao2019} & $\checkmark$ & \cellcolor{gray!50}73.2 & \cellcolor{gray!50}- & \cellcolor{gray!50}- & \cellcolor{gray!50}- \\
    Equivariant3D~\cite{Spezialetti2019} & $\checkmark$ & \cellcolor{gray!50}78.1 & \cellcolor{gray!50}- & \cellcolor{gray!50}- & \cellcolor{gray!50}- \\
    \midrule[0.8pt]
    \multicolumn{6}{l}{\textbf{\emph{Zero-shot methods}}} \\
    \cdashline{1-6}[2pt/1pt] 
    FreeReg~\cite{wang2023freereg} & $\times$ & \cellcolor{gray!50}41.2 & \cellcolor{gray!50}- & \cellcolor{gray!50}15.1 & \cellcolor{gray!50}- \\ 
    \name (Ours) & $\times$ & \cellcolor{gray!50}87.2 & \cellcolor{gray!50}53.9 & \cellcolor{gray!50}52.1 & \cellcolor{gray!50}26.2 \\
    \bottomrule[1pt]
\end{tabular}%
}
\end{table}

\mypara{Comparison of Handcrafted and Zero-Shot PCR Approaches.}
Our approach significantly outperforms handcrafted methods. 
This suggests that the semantic features transferred from foundation models are meaningful.
While FreeReg~\cite{wang2023freereg} enhances image-to-point cloud registration using a pretrained diffusion model in a zero-shot manner, it lacks semantic guidance. 
In addition, FreeReg relies solely on distinctive features, which are limited in capturing spatial relationships within the point cloud. In contrast, our method leverages object matching to achieve higher Registration Recall (RR) than FreeReg, with 46.0\% on 3DMatch and 37.0\% on 3DLoMatch. 
By effectively capturing spatial relationships among various objects within a single point cloud, our approach enables accurate object matching based on semantic features.

\mypara{Comparison of Unsupervised PCR Approaches.}
For unsupervised methods, our approach significantly outperforms current methods on both 3DMatch and 3DLoMatch. Compared to Equivariant3D, our approach achieves a 9.1\% improvement in Registration Recall (RR) on 3DMatch, as shown in ~\cref{exp:tab:3d_3dl}. While unsupervised learning reduces the need for labeled data, the absence of annotated data often leads to suboptimal performance, as these methods rely solely on learned 3D features for registration, without leveraging semantic information.
In contrast, our approach utilizes robust semantic features from foundation models for object matching, effectively capturing both spatial relationships and contextual meanings of objects within the scene. 
This semantic understanding not only improves alignment accuracy but also enhances resilience against noise, resulting in more reliable and precise point cloud registration.

\begin{figure*}[!t]
\centering
\resizebox{\textwidth}{!}{\includegraphics{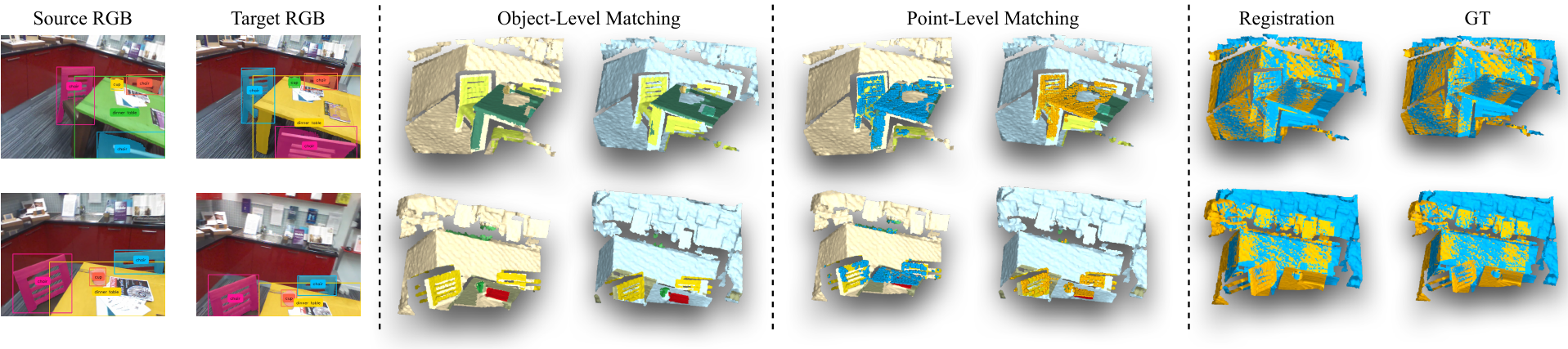}}
\caption{Visualization of the entire process in \name on 3DMatch.
For object-level matching, the matched regions are highlighted using the same color.
For point-level matching with yellow and blue points \textcolor{cusblue}{blue} and \textcolor{cusyellow}{yellow} points are used to visualize the correspondences.
}
\label{exp:visualization}
\vspace{-2mm}
\end{figure*}

\mypara{Comparison of Supervised PCR Approaches.}
Among the evaluated baselines in ~\cref{exp:tab:3d_3dl}, \name achieves performance comparable to most supervised methods on 3DMatch and 3DLoMatch, as shown in \cref{exp:tab:3d_3dl}. While our zero-shot approach doesn’t match the latest SOTA supervised methods on 3D data, it yields results similar to earlier supervised approaches. Compared to CoFiNet’s 88.4\% RR, which employs a coarse-to-fine strategy, our approach achieves comparable performance with 87.2\% RR on 3DMatch. 
This further demonstrates the effectiveness of our object-to-point framework, which is also coarse-to-fine by analogy.
Notably, a main advantage of \name is that it does not require training on 3D labeled data, offering improved adaptability and scalability in data-scarce scenarios.
Although our method still falls short compared to supervised registration methods such as GeoTransformer~\cite{qin2023geotransformer}, which are trained and tested on 3DMatch, this result illustrates the potential for zero-shot PCR. 
Our method’s robustness and adaptability allow it to perform effectively without any 3D annotated training data, enabling strong generalization across diverse scenarios.
~\cref{exp:visualization} visualize the whole registration process of \name on 3DMatch.
More visualization results are provided in the Appendix.

\subsection{Comparison on ScanNet}
\label{exp:comparisons_scan}
We also conduct experiments on ScanNet, with results reported in \cref{exp:tab:scannet}. \name achieves comparable performance in both rotation and translation alignment to FCGF, demonstrating the effectiveness of transferring semantic features from foundation models for object matching.
For unsupervised methods, 
(1) our approach falls short in registration accuracy, particularly at the 5° and 10° thresholds. 
However, at the 45° and 25 cm thresholds, our method demonstrates comparable performance, indicating that it can achieve similar precision at broader alignment tolerances.
(2) Regarding cross-dataset generalization, all unsupervised methods show performance degradation. 
For instance, BYOC, when trained on ScanNet, experiences a 20.0\% drop in accuracy at the 5° threshold and a 10.0\% drop at the 10° threshold when applied to 3DMatch.
This illustrates that \name, by leveraging features transferred from foundation models, achieves improved robustness and generalization across datasets. \name performs well even in data-scarce cases, reducing the need for labeled 3D data.
\begin{table}[t]
\centering
\caption{\textbf{Pairwise Registration on ScanNet.} 
We compare our approach with the supervised geometric method FCGF trained on 3DMatch with pose supervision, and the unsupervised learning method BYOC trained on 3DMatch without pose supervision. 
\textit{P.S} indicates pose supervision. \textit{T.S} refers to the training dataset, where 3D=3DMatch and SC=ScanNet. 
Our approach achieves competitive performance on ScanNet, demonstrating the effectiveness of our zero-shot registration method.}
\label{exp:tab:scannet}
\vspace{-2mm}
\setlength{\tabcolsep}{2pt} 
\resizebox{\columnwidth}{!}{
    \begin{tabular}{l cc  ccccc ccccc ccccc}
    \toprule[0.1em]
    & & & \multicolumn{5}{c}{Rotation} & \multicolumn{5}{c}{Translation} \\
    
    \textbf{Method} & T.S & P.s  
    & \multicolumn{3}{c}{Acc $\uparrow$} & \multicolumn{2}{c}{Err $\downarrow$} 
    & \multicolumn{3}{c}{Acc $\uparrow$} & \multicolumn{2}{c}{Err $\downarrow$} 
    \\ 
    \cmidrule(lr){4-6} \cmidrule(lr){7-8} \cmidrule(lr){9-11} \cmidrule(lr){12-13}
    & & & $5^\circ$ & $10^\circ$ & $45^\circ$ & M & Med. 
    & 5 & 10 & 25 & M & Med. 
    \\
    \midrule
    \multicolumn{7}{l}{\textbf{\emph{Supervised geometric method}}} \\
    \cdashline{1-13}[2pt/1pt] 
    FCGF~\cite{choy2019fully} & 3D & \checkmark  &
    \cellcolor{gray!50}70.2 & \cellcolor{gray!50}87.7 & \cellcolor{gray!50}96.6 & \cellcolor{gray!50}9.5 & \cellcolor{gray!50}3.3 & \cellcolor{gray!50}27.5 & \cellcolor{gray!50}58.3 & \cellcolor{gray!50}23.6 & \cellcolor{gray!50}8.3 & \cellcolor{gray!50}8.3  \\
    \midrule
    \multicolumn{7}{l}{\textbf{\emph{Unsupervised learning-based methods}}} \\
    \cdashline{1-13}[2pt/1pt] 
    BYOC~\cite{banani2021bootstrap} & 3D & $\times$  & 
    \cellcolor{gray!50}66.5 & \cellcolor{gray!50}85.2 & \cellcolor{gray!50}97.8 & \cellcolor{gray!50}7.4 & \cellcolor{gray!50}3.3 & \cellcolor{gray!50}30.7 & \cellcolor{gray!50}57.6 & \cellcolor{gray!50}88.9 & \cellcolor{gray!50}16.0 & \cellcolor{gray!50}8.2 \\
    UR\&R~\cite{mei2023unsupervised} & 3D & $\times$  &  
    \cellcolor{gray!50}87.6 & \cellcolor{gray!50}93.1 & \cellcolor{gray!50}98.3 & \cellcolor{gray!50}4.3 & \cellcolor{gray!50}1.0 & \cellcolor{gray!50}69.2 & \cellcolor{gray!50}84.0 & \cellcolor{gray!50}93.8 & \cellcolor{gray!50}9.5  & \cellcolor{gray!50}2.8  \\
    BYOC~\cite{banani2021bootstrap} & SC & $\times$  & 
    \cellcolor{gray!50}86.5 & \cellcolor{gray!50}95.2 & \cellcolor{gray!50}\textbf{99.1} & \cellcolor{gray!50}\textbf{3.8} & \cellcolor{gray!50}1.7 & \cellcolor{gray!50}56.4 & \cellcolor{gray!50}80.6 & \cellcolor{gray!50}\textbf{96.3} & \cellcolor{gray!50}8.7 & \cellcolor{gray!50}4.3 \\
    UR\&R~\cite{mei2023unsupervised} & SC & $\times$  & 
    \cellcolor{gray!50}\textbf{92.7} & \cellcolor{gray!50}\textbf{95.8} & \cellcolor{gray!50}{98.5} & \cellcolor{gray!50}3.4 & \cellcolor{gray!50}\textbf{0.8} & \cellcolor{gray!50}\textbf{77.2} & \cellcolor{gray!50}\textbf{89.6} & \cellcolor{gray!50}{96.1} & \cellcolor{gray!50}\textbf{7.3}& \cellcolor{gray!50}\textbf{2.3}  \\
    
    \midrule
    \multicolumn{7}{l}{\textbf{\emph{Zero-shot method}}} \\
    \cdashline{1-13}[2pt/1pt] 
    \name (ours) & $ \times$ & $\times$ & 
    \cellcolor{gray!50}71.2 & \cellcolor{gray!50}87.3 & \cellcolor{gray!50}97.7 & \cellcolor{gray!50}8.0 & \cellcolor{gray!50}3.1 & \cellcolor{gray!50}23.5 & \cellcolor{gray!50}60.2 & \cellcolor{gray!50}85.3 & \cellcolor{gray!50}15.2  & \cellcolor{gray!50}8.9 \\

    \bottomrule[0.1em]
    \end{tabular}
}
\vspace{-2mm}
\end{table}

\subsection{Ablation Studies \& Analysis}
\label{exp:ablation}
We conduct extensive ablation studies to better understand the different modules in \name as shown in \cref{exp:tab:ablation_zeroreg}.

\mypara{Analysis of Model Components.}
\begin{table}[t]
\centering
\caption{Ablation study of ZeroReg components on the 3DMatch dataset. (a) presents the impact of excluding specific Object-Level (Detection, Segmentation, Scene Graph, and CLIP) and Point-Level modules on Registration Recall (RR) and Inlier Ratio (IR), while (b) compares various foundation model configurations for Detection and Segmentation and their effects on RR and IR.}
\label{exp:tab:ablation_zeroreg}
\setlength{\tabcolsep}{4pt} 
\renewcommand{\arraystretch}{1.3} 

\begin{subtable}[t]{0.48\textwidth}
\centering
\tabcolsep 1pt
\caption{Impact of Object-Level and Point-Level modules on RR and IR.}
\label{exp:tab:ablation_zeroreg_a}
\resizebox{\columnwidth}{!}{%
\begin{tabular}{lcccccccc}
\toprule
\multirow{2}{*}{\textbf{Configurations}} & \multicolumn{4}{c}{\textbf{Object-Level}} & \multirow{2}{*}{\textbf{Point-Level}} & \multicolumn{2}{c}{\textbf{3DMatch}} \\
 & Det & Seg & Graph & CLIP & & \textbf{RR(\%)} & \textbf{IR(\%)} \\
\midrule
Full (ZeroReg) & \checkmark & \checkmark & \checkmark & \checkmark & LoFTR & \cellcolor{gray!50}\textbf{87.2} & \cellcolor{gray!50}\textbf{53.9} \\
(a) w/o Object-Level &  \(\times\) & \(\times\) & \(\times\) & \(\times\) & LoFTR & \cellcolor{gray!50}73.3(\red{-13.9\%}) & \cellcolor{gray!50}50.2(\red{-3.7\%}) \\
(b) w/o Point-Level & \checkmark & \checkmark & \checkmark & \checkmark & \(\times\) & \cellcolor{gray!50}71.3(\red{-15.9\%}) & \cellcolor{gray!50}- \\
(c) w/o Detection & \(\times\) & \checkmark & \checkmark & \checkmark & LoFTR & \cellcolor{gray!50}78.5(\red{-8.7\%}) & \cellcolor{gray!50}50.9(\red{-3.0\%})\\
(d) w/o Segmentation & \checkmark & \(\times\) & \checkmark & \checkmark & LoFTR & \cellcolor{gray!50}85.1(\red{-2.2\%}) & \cellcolor{gray!50}52.8(\red{-1.1\%})\\
(e) w/o Scene Graph & \checkmark & \checkmark & \(\times\) & \checkmark & LoFTR & \cellcolor{gray!50}83.2(\red{-4.0\%}) & \cellcolor{gray!50}51.8(\red{-2.1\%})\\
(f) w/o CLIP & \checkmark & \checkmark & \checkmark & \(\times\) & LoFTR & \cellcolor{gray!50}84.5(\red{-2.7\%})& \cellcolor{gray!50}52.0(\red{-1.9\%})\\
(g) Point-Level w/ SG & \checkmark & \checkmark & \checkmark & \checkmark & SuperGlue & \cellcolor{gray!50}81.6(\red{-5.6\%}) & \cellcolor{gray!50}51.2(\red{-2.7\%}) \\
\bottomrule
\end{tabular}%
}
\end{subtable}%
\hfill
\begin{subtable}[t]{0.48\textwidth}
\centering
\caption{Foundation model comparison for Detection and Segmentation.}
\label{exp:tab:ablation_zeroreg_b}
\resizebox{\columnwidth}{!}{%
\begin{tabular}{lcccc}
\toprule
\textbf{Method} & \textbf{Detection Model} & \textbf{Segmentation Model} & \textbf{RR(\%)} & \textbf{IR(\%)} \\
\midrule
(i) \name  & YOLOv10~\cite{wang2024yolov10} & SAMv1~\cite{kirillov2023segment} & \cellcolor{gray!50}78.4 & \cellcolor{gray!50}46.9 \\
(j) \name  & YOLOv10~\cite{wang2024yolov10} & SAMv2~\cite{ravi2024sam} & \cellcolor{gray!50}79.2 & \cellcolor{gray!50}47.4 \\
(k) \name  & Florence-2~\cite{xiao2024florence} & SAMv1~\cite{kirillov2023segment} & \cellcolor{gray!50}85.7 & \cellcolor{gray!50}53.3 \\
Full (\name)  & Florence-2~\cite{xiao2024florence} & SAMv2~\cite{ravi2024sam} & \cellcolor{gray!50}\textbf{87.2} & \cellcolor{gray!50}\textbf{53.9} \\
\bottomrule
\end{tabular}%
}
\end{subtable}
\vspace{-3mm}
\end{table}
We compare seven configurations to evaluate the effectiveness of each module based on \textbf{Registration Recall (RR)} and \textbf{Inlier Ratio (IR)} in \cref{exp:tab:ablation_zeroreg_a}: (a): only Point-Level module; (b): only Object-Level module;(c): based on (b) without Detection;
(d): based on (b) without Segmentation;
(e): based on (b) without Scene Graph;
(f): based on (b) without CLIP, and 
(g) based on (a) but replaced with SuperGlue.

\begin{enumerate}
    \item \textbf{Impact of Object-Level and Point-Level Modules}:
    \begin{itemize}
        \item The \cref{exp:tab:ablation_zeroreg_a} compares the effects of excluding Object-Level (Detection, Segmentation, Scene Graph, and CLIP) and Point-Level modules. Results show that removing the Object-Level modules (Row a) lowers RR to 73.3\% and IR to 50.2\%, while removing the Point-Level module (Row b) significantly reduces RR and IR to 71.3\% and 50.9\%.
        \item This indicates that both Object-Level and Point-Level modules play a crucial role in the overall registration performance. 
        The Object-Level modules provide localization, semantic, and structural information, while the Point-Level module refines alignment. 
        They complement each other to enhance overall accuracy and robustness, confirming that both levels are integral to effective registration.
    \end{itemize}
    \item \textbf{Impact of Detection, Segmentation, Scene Graph, and CLIP Sub-Modules}:
    \noindent \textbf{Detection}: Removing it decreases RR by 10.0\% and IR by 3.7\%, demonstrating its role in initial localization; this highlights the critical importance of detection for accurate initial localization. 
    \textbf{Segmentation}: Excluding it lowers RR by 2.4\% and IR by 1.1\%, indicating its role in refining boundaries; this shows that segmentation aids in enhancing boundary precision for better registration.
    \textbf{Scene Graph}: Its removal reduces RR by 4.6\% and IR by 2.1\%, emphasizing its role in modeling relationships; this confirms that the scene graph is essential for understanding inter-object relationships.
    \textbf{CLIP}: Removing it results in a 3.1\% drop in RR and 1.9\% drop in IR, showing its contribution to semantic matching; this underscores CLIP's value in enhancing semantic alignment within the model.
\end{enumerate}
\mypara{Conbinations of Different Models.}
We compares the specialized detection model YOLOv10~\cite{wang2024yolov10} with the foundation detection model Florence-2~\cite{xiao2024florence}, along with different segmentation models, including SAMv1~\cite{kirillov2023segment} and SAMv2~\cite{ravi2024sam} as shown in~\cref{exp:tab:ablation_zeroreg_b}.
\begin{itemize}
    \item \textbf{Task-Specific vs Foundation Models}: Pairing YOLOv10 with SAMv1 or SAMv2 yields moderate RR values (78.4\% to 79.2\%), indicating that task-specific models, trained on a limited set of categories, are effective for specific tasks but may not generalize across diverse applications as well as foundation models.
    \item \textbf{Segmentation Comparison}: With precise bounding boxes as prompt inputs, SAMv1 and SAMv2 perform equally well in segmentation, with comparable results.
\end{itemize}

\mypara{Impact of Number of Views.} As shown in ~\cref{exp:tab:num_views}, increasing the number of views improves registration recall, though it generally comes at the cost of reduced efficiency.
More ablation studies are provided in the Appendix.
\begin{table}[t]
\centering
\small
\tabcolsep 11pt
\caption{Ablation study on impact of the number of views.}
\label{exp:tab:num_views}
\setlength{\tabcolsep}{4pt} 
\renewcommand{\arraystretch}{0.8} 
\resizebox{0.85\columnwidth}{!}
{%
\begin{tabular}{lccccc}
\toprule
\multirow{2}{*}{\textbf{Model}} & \multirow{2}{*}{\textbf{Views}} & \multicolumn{2}{c}{\textbf{3DMatch}} & \multicolumn{2}{c}{\textbf{3DLoMatch}} \\
 & & \textbf{RR(\%)} & \textbf{IR(\%)} & \textbf{RR(\%)} & \textbf{IR(\%)} \\
\midrule
\name & 1 & \cellcolor{gray!50}85.2 & \cellcolor{gray!50}51.9 & \cellcolor{gray!50}50.1 & \cellcolor{gray!50}23.9 \\
\name & 2 & \cellcolor{gray!50}87.0 & \cellcolor{gray!50}53.5 & \cellcolor{gray!50}51.8 & \cellcolor{gray!50}25.6\\
\name & 3 & \cellcolor{gray!50}\textbf{87.2} & \cellcolor{gray!50}\textbf{53.9} & \cellcolor{gray!50}\textbf{52.1} & \cellcolor{gray!50}\textbf{26.2} \\
\bottomrule
\end{tabular}%
}
\end{table}

\section{Conclusion}
In this paper, we propose a zero-shot method for point cloud registration that identifies matched objects based on semantic similarity, removing the need for additional 3D data training. First, we leverage foundation models to localize object positions through detection and segmentation. Next, we construct scene graphs centered on the centroids of different objects, capturing their contextual positional relationships to address semantic ambiguity. Using reliably matched objects, we then predict point-level correspondences within the matched regions of the source and target. Finally, we complete the registration process.
We conducted extensive experiments, demonstrating that the transferred semantic features are meaningful for registration.
Especially as our approach requires no training on 3D data, it is highly scalable for scenarios where data is limited.
\vspace{-2mm}
\paragraph{Limitations}
Foundation models are primarily trained on 2D images, which lack an intrinsic understanding of 3D scenes. 
This creates a modality gap when transferring from 2D to 3D—a key limitation impacting zero-shot PCR performance and requires extra adjustments to handle the modality gap between 2D and 3D data.



{
    \small
    \bibliographystyle{ieeenat_fullname}
    \bibliography{main}

\begin{thebibliography}{66}
\providecommand{\natexlab}[1]{#1}
\providecommand{\url}[1]{\texttt{#1}}
\expandafter\ifx\csname urlstyle\endcsname\relax
  \providecommand{\doi}[1]{doi: #1}\else
  \providecommand{\doi}{doi: \begingroup \urlstyle{rm}\Url}\fi

\bibitem[Aiger et~al.(2008)Aiger, Mitra, and Cohen-Or]{aiger20084}
Dror Aiger, Niloy~J Mitra, and Daniel Cohen-Or.
\newblock 4-points congruent sets for robust pairwise surface registration.
\newblock In \emph{ACM SIGGRAPH}, pages 1--10. 2008.

\bibitem[Ao et~al.(2021)Ao, Hu, Yang, Markham, and Guo]{ao2020SpinNet}
Sheng Ao, Qingyong Hu, Bo Yang, Andrew Markham, and Yulan Guo.
\newblock Spinnet: Learning a general surface descriptor for 3d point cloud registration.
\newblock In \emph{CVPR}, 2021.

\bibitem[Bai et~al.(2020)Bai, Luo, Zhou, Fu, Quan, and Tai]{bai2020d3feat}
Xuyang Bai, Zixin Luo, Lei Zhou, Hongbo Fu, Long Quan, and Chiew-Lan Tai.
\newblock D3feat: Joint learning of dense detection and description of 3d local features.
\newblock In \emph{CVPR}, pages 6359--6367, 2020.

\bibitem[Bay et~al.(2006)Bay, Tuytelaars, and Van~Gool]{bay2006surf}
Herbert Bay, Tinne Tuytelaars, and Luc Van~Gool.
\newblock Surf: Speeded up robust features.
\newblock In \emph{Computer Vision--ECCV 2006: 9th European Conference on Computer Vision, Graz, Austria, May 7-13, 2006. Proceedings, Part I 9}, pages 404--417. Springer, 2006.

\bibitem[Besl and McKay(1992)]{Besl1992}
P.J. Besl and N.D. McKay.
\newblock {A method for registration of 3-D shapes}.
\newblock \emph{PAMI}, 14\penalty0 (2):\penalty0 239--256, 1992.

\bibitem[Busam et~al.(2015)Busam, Esposito, Che'Rose, Navab, and Frisch]{7406423}
Benjamin Busam, Marco Esposito, Simon Che'Rose, Nassir Navab, and Benjamin Frisch.
\newblock A stereo vision approach for cooperative robotic movement therapy.
\newblock In \emph{ICCV workshop}, pages 519--527, 2015.

\bibitem[Choy et~al.(2019)Choy, Park, and Koltun]{choy2019fully}
Christopher Choy, Jaesik Park, and Vladlen Koltun.
\newblock Fully convolutional geometric features.
\newblock In \emph{ICCV}, pages 8958--8966, 2019.

\bibitem[Cirujeda et~al.(2014)Cirujeda, Mateo, Dicente, and Binefa]{cirujeda2014mcov}
Pol Cirujeda, Xavier Mateo, Yashin Dicente, and Xavier Binefa.
\newblock Mcov: a covariance descriptor for fusion of texture and shape features in 3d point clouds.
\newblock In \emph{3DV}, pages 551--558. IEEE, 2014.

\bibitem[Cuturi(2013)]{cuturi2013sinkhorn}
Marco Cuturi.
\newblock Sinkhorn distances: Lightspeed computation of optimal transport.
\newblock \emph{NeuriPS}, 26, 2013.

\bibitem[Dai et~al.(2017)Dai, Chang, Savva, Halber, Funkhouser, and Nie{\ss}ner]{dai2017scannet}
Angela Dai, Angel~X. Chang, Manolis Savva, Maciej Halber, Thomas Funkhouser, and Matthias Nie{\ss}ner.
\newblock Scannet: Richly-annotated 3d reconstructions of indoor scenes.
\newblock In \emph{CVPR}, 2017.

\bibitem[Deng et~al.(2018{\natexlab{a}})Deng, Birdal, and Ilic]{Deng2018eccv}
H. Deng, T. Birdal, and S. Ilic.
\newblock {PPF-FoldNet: Unsupervised learning of rotation invariant 3D local descriptors}.
\newblock In \emph{ECCV}, 2018{\natexlab{a}}.

\bibitem[Deng et~al.(2018{\natexlab{b}})Deng, Birdal, and Ilic]{deng2018ppfnet}
Haowen Deng, Tolga Birdal, and Slobodan Ilic.
\newblock Ppfnet: Global context aware local features for robust 3d point matching.
\newblock In \emph{CVPR}, pages 195--205, 2018{\natexlab{b}}.

\bibitem[DeTone et~al.(2018)DeTone, Malisiewicz, and Rabinovich]{detone2018superpoint}
Daniel DeTone, Tomasz Malisiewicz, and Andrew Rabinovich.
\newblock Superpoint: Self-supervised interest point detection and description.
\newblock In \emph{CVPR}, pages 224--236, 2018.

\bibitem[Drost et~al.(2010)Drost, Ulrich, Navab, and Ilic]{drost2010model}
Bertram Drost, Markus Ulrich, Nassir Navab, and Slobodan Ilic.
\newblock Model globally, match locally: Efficient and robust 3d object recognition.
\newblock In \emph{CVPR}, pages 998--1005. Ieee, 2010.

\bibitem[El~Banani and Johnson(2021)]{banani2021bootstrap}
Mohamed El~Banani and Justin Johnson.
\newblock Bootstrap your own correspondences.
\newblock In \emph{ICCV}, pages 6433--6442, 2021.

\bibitem[El~Banani and Johnson(2022)]{wang2022improving}
Mohamed El~Banani and Justin Johnson.
\newblock Improving rgb-d point cloud registration by learning multi-scale local linear transformation.
\newblock In \emph{ECCV}, 2022.

\bibitem[El~Banani et~al.(2021)El~Banani, Gao, and Johnson]{banani2021unsupervisedrr}
Mohamed El~Banani, Luya Gao, and Justin Johnson.
\newblock Unsupervisedr\&r: Unsupervised point cloud registration via differentiable rendering.
\newblock In \emph{CVPR}, pages 7129--7139, 2021.

\bibitem[Fischler and Bolles(1981)]{fischler1981random}
Martin~A Fischler and Robert~C Bolles.
\newblock Random sample consensus: a paradigm for model fitting with applications to image analysis and automated cartography.
\newblock \emph{Communications of the ACM}, 24\penalty0 (6):\penalty0 381--395, 1981.

\bibitem[Gojcic et~al.(2019)Gojcic, Zhou, Wegner, and Wieser]{gojcic2019perfect}
Zan Gojcic, Caifa Zhou, Jan~D Wegner, and Andreas Wieser.
\newblock The perfect match: 3d point cloud matching with smoothed densities.
\newblock In \emph{CVPR}, pages 5545--5554, 2019.

\bibitem[Hatem et~al.(2023)Hatem, Qian, and Wang]{Hatem_2023_ICCV}
Ahmed Hatem, Yiming Qian, and Yang Wang.
\newblock Point-tta: Test-time adaptation for point cloud registration using multitask meta-auxiliary learning.
\newblock In \emph{ICCV}, pages 16494--16504, 2023.

\bibitem[Huang et~al.(2021)Huang, Gojcic, Usvyatsov, Wieser, and Schindler]{huang2021predator}
Shengyu Huang, Zan Gojcic, Mikhail Usvyatsov, Andreas Wieser, and Konrad Schindler.
\newblock Predator: Registration of 3d point clouds with low overlap.
\newblock In \emph{CVPR}, pages 4267--4276, 2021.

\bibitem[Huang et~al.(2020)Huang, Mei, and Zhang]{huang2020feature}
Xiaoshui Huang, Guofeng Mei, and Jian Zhang.
\newblock Feature-metric registration: A fast semi-supervised approach for robust point cloud registration without correspondences.
\newblock In \emph{CVPR}, pages 11366--11374, 2020.

\bibitem[Ilharco et~al.(2021)Ilharco, Wortsman, Wightman, Gordon, Carlini, Taori, Dave, Shankar, Namkoong, Miller, Hajishirzi, Farhadi, and Schmidt]{openclip}
Gabriel Ilharco, Mitchell Wortsman, Ross Wightman, Cade Gordon, Nicholas Carlini, Rohan Taori, Achal Dave, Vaishaal Shankar, Hongseok Namkoong, John Miller, Hannaneh Hajishirzi, Ali Farhadi, and Ludwig Schmidt.
\newblock Openclip, 2021.
\newblock If you use this software, please cite it as below.

\bibitem[Jatavallabhula et~al.(2023)Jatavallabhula, Kuwajerwala, Gu, Omama, Chen, Maalouf, Li, Iyer, Saryazdi, Keetha, et~al.]{jatavallabhula2023conceptfusion}
Krishna~Murthy Jatavallabhula, Alihusein Kuwajerwala, Qiao Gu, Mohd Omama, Tao Chen, Alaa Maalouf, Shuang Li, Ganesh Iyer, Soroush Saryazdi, Nikhil Keetha, et~al.
\newblock Conceptfusion: Open-set multimodal 3d mapping.
\newblock \emph{arXiv preprint arXiv:2302.07241}, 2023.

\bibitem[Jia et~al.(2021)Jia, Yang, Xia, Chen, Parekh, Pham, Le, Sung, Li, and Duerig]{jia2021scaling}
Chao Jia, Yinfei Yang, Ye Xia, Yi-Ting Chen, Zarana Parekh, Hieu Pham, Quoc Le, Yun-Hsuan Sung, Zhen Li, and Tom Duerig.
\newblock Scaling up visual and vision-language representation learning with noisy text supervision.
\newblock In \emph{ICML}, pages 4904--4916. PMLR, 2021.

\bibitem[Johnson and Hebert(1999)]{johnson1999using}
Andrew~E Johnson and Martial Hebert.
\newblock Using spin images for efficient object recognition in cluttered 3d scenes.
\newblock \emph{IEEE TPAMI}, 21\penalty0 (5):\penalty0 433--449, 1999.

\bibitem[Kang et~al.(2024)Kang, Luan, Khoshelham, and Wang]{Kang_2024}
Xueyang Kang, Zhaoliang Luan, Kourosh Khoshelham, and Bing Wang.
\newblock \emph{Equi-GSPR: Equivariant SE(3) Graph Network Model for Sparse Point Cloud Registration}, page 149–167.
\newblock Springer Nature Switzerland, 2024.

\bibitem[Kirillov et~al.(2023)Kirillov, Mintun, Ravi, Mao, Rolland, Gustafson, Xiao, Whitehead, Berg, Lo, et~al.]{kirillov2023segment}
Alexander Kirillov, Eric Mintun, Nikhila Ravi, Hanzi Mao, Chloe Rolland, Laura Gustafson, Tete Xiao, Spencer Whitehead, Alexander~C Berg, Wan-Yen Lo, et~al.
\newblock Segment anything.
\newblock In \emph{CVPR}, pages 4015--4026, 2023.

\bibitem[Li et~al.(2023)Li, Li, Savarese, and Hoi]{li2023blip}
Junnan Li, Dongxu Li, Silvio Savarese, and Steven Hoi.
\newblock Blip-2: Bootstrapping language-image pre-training with frozen image encoders and large language models.
\newblock \emph{arXiv preprint arXiv:2301.12597}, 2023.

\bibitem[Liu et~al.(2023{\natexlab{a}})Liu, Wang, Liu, Jiang, Pollefeys, and Wang]{Liu_2023_ICCV}
Jiuming Liu, Guangming Wang, Zhe Liu, Chaokang Jiang, Marc Pollefeys, and Hesheng Wang.
\newblock Regformer: An efficient projection-aware transformer network for large-scale point cloud registration.
\newblock In \emph{ICCV}, pages 8451--8460, 2023{\natexlab{a}}.

\bibitem[Liu et~al.(2023{\natexlab{b}})Liu, Zhu, Zhou, Li, Chang, and Guo]{Liu1_2023_ICCV}
Quan Liu, Hongzi Zhu, Yunsong Zhou, Hongyang Li, Shan Chang, and Minyi Guo.
\newblock Density-invariant features for distant point cloud registration.
\newblock In \emph{ICCV}, pages 18215--18225, 2023{\natexlab{b}}.

\bibitem[Loiola et~al.(2007)Loiola, De~Abreu, Boaventura-Netto, Hahn, and Querido]{loiola2007survey}
Eliane~Maria Loiola, Nair Maria~Maia De~Abreu, Paulo~Oswaldo Boaventura-Netto, Peter Hahn, and Tania Querido.
\newblock A survey for the quadratic assignment problem.
\newblock \emph{European journal of operational research}, 176\penalty0 (2):\penalty0 657--690, 2007.

\bibitem[Lowe(2004)]{lowe2004distinctive}
David~G Lowe.
\newblock Distinctive image features from scale-invariant keypoints.
\newblock \emph{IJCV}, 60:\penalty0 91--110, 2004.

\bibitem[Ma et~al.(2024)Ma, Li, Ren, Sebe, Konukoglu, Gevers, Van~Gool, and Paudel]{ma2024shapesplat}
Qi Ma, Yue Li, Bin Ren, Nicu Sebe, Ender Konukoglu, Theo Gevers, Luc Van~Gool, and Danda~Pani Paudel.
\newblock Shapesplat: A large-scale dataset of gaussian splats and their self-supervised pretraining.
\newblock In \emph{3DV}, 2024.

\bibitem[Mei et~al.(2023{\natexlab{a}})Mei, Poiesi, Saltori, Zhang, Ricci, and Sebe]{mei2023overlap}
Guofeng Mei, Fabio Poiesi, Cristiano Saltori, Jian Zhang, Elisa Ricci, and Nicu Sebe.
\newblock Overlap-guided gaussian mixture models for point cloud registration.
\newblock In \emph{WACV}, pages 4511--4520, 2023{\natexlab{a}}.

\bibitem[Mei et~al.(2023{\natexlab{b}})Mei, Tang, Huang, Wang, Liu, Zhang, Van~Gool, and Wu]{mei2023unsupervised}
Guofeng Mei, Hao Tang, Xiaoshui Huang, Weijie Wang, Juan Liu, Jian Zhang, Luc Van~Gool, and Qiang Wu.
\newblock Unsupervised deep probabilistic approach for partial point cloud registration.
\newblock In \emph{CVPR}, pages 13611--13620, 2023{\natexlab{b}}.

\bibitem[Moravec(1981)]{moravec1981rover}
Hans~P Moravec.
\newblock Rover visual obstacle avoidance.
\newblock In \emph{IJCAI}, pages 785--790, 1981.

\bibitem[Paszke et~al.(2019)Paszke, Gross, Massa, Lerer, Bradbury, Chanan, Killeen, Lin, Gimelshein, Antiga, et~al.]{paszke2019pytorch}
Adam Paszke, Sam Gross, Francisco Massa, Adam Lerer, James Bradbury, Gregory Chanan, Trevor Killeen, Zeming Lin, Natalia Gimelshein, Luca Antiga, et~al.
\newblock Pytorch: An imperative style, high-performance deep learning library.
\newblock \emph{NeuriPS}, 32, 2019.

\bibitem[Peng et~al.(2023)Peng, Genova, Jiang, Tagliasacchi, Pollefeys, and Funkhouser]{peng2023openscene}
Songyou Peng, Kyle Genova, Chiyu~"Max" Jiang, Andrea Tagliasacchi, Marc Pollefeys, and Thomas Funkhouser.
\newblock Openscene: 3d scene understanding with open vocabularies, 2023.

\bibitem[Qin et~al.(2023)Qin, Yu, Wang, Guo, Peng, Ilic, Hu, and Xu]{qin2023geotransformer}
Zheng Qin, Hao Yu, Changjian Wang, Yulan Guo, Yuxing Peng, Slobodan Ilic, Dewen Hu, and Kai Xu.
\newblock Geotransformer: Fast and robust point cloud registration with geometric transformer.
\newblock \emph{IEEE TPAMI}, 2023.

\bibitem[Radford et~al.(2021)Radford, Kim, Hallacy, Ramesh, Goh, Agarwal, Sastry, Askell, Mishkin, Clark, et~al.]{radford2021learning}
Alec Radford, Jong~Wook Kim, Chris Hallacy, Aditya Ramesh, Gabriel Goh, Sandhini Agarwal, Girish Sastry, Amanda Askell, Pamela Mishkin, Jack Clark, et~al.
\newblock Learning transferable visual models from natural language supervision.
\newblock In \emph{ICML}, pages 8748--8763. PMLR, 2021.

\bibitem[Ravi et~al.(2024)Ravi, Gabeur, Hu, Hu, Ryali, Ma, Khedr, R{\"a}dle, Rolland, Gustafson, et~al.]{ravi2024sam}
Nikhila Ravi, Valentin Gabeur, Yuan-Ting Hu, Ronghang Hu, Chaitanya Ryali, Tengyu Ma, Haitham Khedr, Roman R{\"a}dle, Chloe Rolland, Laura Gustafson, et~al.
\newblock Sam 2: Segment anything in images and videos.
\newblock \emph{arXiv preprint arXiv:2408.00714}, 2024.

\bibitem[Ren et~al.(2024)Ren, Mei, Paudel, Wang, Li, Liu, Cucchiara, Van~Gool, and Sebe]{ren2024bringing}
Bin Ren, Guofeng Mei, Danda~Pani Paudel, Weijie Wang, Yawei Li, Mengyuan Liu, Rita Cucchiara, Luc Van~Gool, and Nicu Sebe.
\newblock Bringing masked autoencoders explicit contrastive properties for point cloud self-supervised learning.
\newblock In \emph{ACCV}, 2024.

\bibitem[Rublee et~al.(2011)Rublee, Rabaud, Konolige, and Bradski]{rublee2011orb}
Ethan Rublee, Vincent Rabaud, Kurt Konolige, and Gary Bradski.
\newblock Orb: An efficient alternative to sift or surf.
\newblock In \emph{2011 International conference on computer vision}, pages 2564--2571. Ieee, 2011.

\bibitem[Rusu et~al.(2008)Rusu, Blodow, Marton, and Beetz]{rusu2008aligning}
Radu~Bogdan Rusu, Nico Blodow, Zoltan~Csaba Marton, and Michael Beetz.
\newblock Aligning point cloud views using persistent feature histograms.
\newblock In \emph{2008 IEEE/RSJ international conference on intelligent robots and systems}, pages 3384--3391. IEEE, 2008.

\bibitem[Rusu et~al.(2009)Rusu, Blodow, and Beetz]{rusu2009fast}
Radu~Bogdan Rusu, Nico Blodow, and Michael Beetz.
\newblock Fast point feature histograms (fpfh) for 3d registration.
\newblock In \emph{ICRA}, pages 3212--3217. IEEE, 2009.

\bibitem[Salti et~al.(2014)Salti, Tombari, and Di~Stefano]{salti2014shot}
Samuele Salti, Federico Tombari, and Luigi Di~Stefano.
\newblock Shot: Unique signatures of histograms for surface and texture description.
\newblock \emph{CVIU}, 125:\penalty0 251--264, 2014.

\bibitem[Sarlin et~al.(2020)Sarlin, DeTone, Malisiewicz, and Rabinovich]{sarlin2020superglue}
Paul-Edouard Sarlin, Daniel DeTone, Tomasz Malisiewicz, and Andrew Rabinovich.
\newblock Superglue: Learning feature matching with graph neural networks.
\newblock In \emph{CVPR}, pages 4938--4947, 2020.

\bibitem[Spezialetti et~al.(2019)Spezialetti, Salti, and Stefano]{Spezialetti2019}
R. Spezialetti, S. Salti, and L.~Di Stefano.
\newblock {Learning an Effective Equivariant 3D Descriptor Without Supervision }.
\newblock In \emph{ICCV}, 2019.

\bibitem[Sun et~al.(2021)Sun, Shen, Wang, Bao, and Zhou]{sun2021loftr}
Jiaming Sun, Zehong Shen, Yuang Wang, Hujun Bao, and Xiaowei Zhou.
\newblock Loftr: Detector-free local feature matching with transformers.
\newblock In \emph{CVPR}, pages 8922--8931, 2021.

\bibitem[Tombari et~al.(2010)Tombari, Salti, and Stefano]{Tombari2010}
F. Tombari, S. Salti, and L.~Di Stefano.
\newblock {Unique shape context for 3D data description}.
\newblock In \emph{ACM 3D Object Retrieval}, 2010.

\bibitem[Wang et~al.(2024{\natexlab{a}})Wang, Chen, Liu, Chen, Lin, Han, and Ding]{wang2024yolov10}
Ao Wang, Hui Chen, Lihao Liu, Kai Chen, Zijia Lin, Jungong Han, and Guiguang Ding.
\newblock Yolov10: Real-time end-to-end object detection.
\newblock \emph{arXiv preprint arXiv:2405.14458}, 2024{\natexlab{a}}.

\bibitem[Wang et~al.(2022)Wang, Liu, Dong, and Wang]{wang2022you}
Haiping Wang, Yuan Liu, Zhen Dong, and Wenping Wang.
\newblock You only hypothesize once: Point cloud registration with rotation-equivariant descriptors.
\newblock In \emph{ACM MM}, 2022.

\bibitem[Wang et~al.(2024{\natexlab{b}})Wang, Liu, Wang, Sun, Dong, Wang, and Yang]{wang2023freereg}
Haiping Wang, Yuan Liu, Bing Wang, Yujing Sun, Zhen Dong, Wenping Wang, and Bisheng Yang.
\newblock Freereg: Image-to-point cloud registration leveraging pretrained diffusion models and monocular depth estimators.
\newblock In \emph{ICLR}, 2024{\natexlab{b}}.

\bibitem[Xiao et~al.(2024)Xiao, Wu, Xu, Dai, Hu, Lu, Zeng, Liu, and Yuan]{xiao2024florence}
Bin Xiao, Haiping Wu, Weijian Xu, Xiyang Dai, Houdong Hu, Yumao Lu, Michael Zeng, Ce Liu, and Lu Yuan.
\newblock Florence-2: Advancing a unified representation for a variety of vision tasks.
\newblock In \emph{CVPR}, pages 4818--4829, 2024.

\bibitem[Yang et~al.(2020)Yang, Shi, and Carlone]{yang2020teaser}
Heng Yang, Jingnan Shi, and Luca Carlone.
\newblock Teaser: Fast and certifiable point cloud registration.
\newblock \emph{IEEE Transactions on Robotics}, 37\penalty0 (2):\penalty0 314--333, 2020.

\bibitem[Yang et~al.(2018)Yang, Feng, Shen, and Tian]{yang2018foldingnet}
Yaoqing Yang, Chen Feng, Yiru Shen, and Dong Tian.
\newblock Foldingnet: Point cloud auto-encoder via deep grid deformation.
\newblock In \emph{Proceedings of the IEEE conference on computer vision and pattern recognition}, pages 206--215, 2018.

\bibitem[Yu et~al.(2021)Yu, Li, Saleh, Busam, and Ilic]{yu2021cofinet}
Hao Yu, Fu Li, Mahdi Saleh, Benjamin Busam, and Slobodan Ilic.
\newblock Cofinet: Reliable coarse-to-fine correspondences for robust point cloud registration.
\newblock In \emph{Neurips}, 2021.

\bibitem[Yu et~al.(2022)Yu, Hou, Qin, Saleh, Shugurov, Wang, Busam, and Ilic]{yu2022riga}
Hao Yu, Ji Hou, Zheng Qin, Mahdi Saleh, Ivan Shugurov, Kai Wang, Benjamin Busam, and Slobodan Ilic.
\newblock Riga: Rotation-invariant and globally-aware descriptors for point cloud registration.
\newblock \emph{arXiv preprint arXiv:2209.13252}, 2022.

\bibitem[Yu et~al.(2023)Yu, Qin, Hou, Saleh, Li, Busam, and Ilic]{yu2023rotationinvariant}
Hao Yu, Zheng Qin, Ji Hou, Mahdi Saleh, Dongsheng Li, Benjamin Busam, and Slobodan Ilic.
\newblock Rotation-invariant transformer for point cloud matching.
\newblock In \emph{CVPR}, 2023.

\bibitem[Zeng et~al.(2017)Zeng, Song, Nie{\ss}ner, Fisher, Xiao, and Funkhouser]{zeng20173dmatch}
Andy Zeng, Shuran Song, Matthias Nie{\ss}ner, Matthew Fisher, Jianxiong Xiao, and Thomas Funkhouser.
\newblock 3dmatch: Learning local geometric descriptors from rgb-d reconstructions.
\newblock In \emph{CVPR}, pages 1802--1811, 2017.

\bibitem[Zhang et~al.(2022{\natexlab{a}})Zhang, Guo, Zhang, Li, Miao, Cui, Qiao, Gao, and Li]{zhang2022pointclip}
Renrui Zhang, Ziyu Guo, Wei Zhang, Kunchang Li, Xupeng Miao, Bin Cui, Yu Qiao, Peng Gao, and Hongsheng Li.
\newblock Pointclip: Point cloud understanding by clip.
\newblock In \emph{CVPR}, pages 8552--8562, 2022{\natexlab{a}}.

\bibitem[Zhang et~al.(2022{\natexlab{b}})Zhang, Yu, Huang, Zhou, and Hou]{zhang2023pcrcg}
Yu Zhang, Junle Yu, Xiaolin Huang, Wenhui Zhou, and Ji Hou.
\newblock Pcr-cg: Point cloud registration via deep explicit color and geometry.
\newblock In \emph{ECCV}, pages 443--459. Springer, 2022{\natexlab{b}}.

\bibitem[Zhao et~al.(2019)Zhao, Birdal, Deng, and Tombari]{Zhao2019}
Y. Zhao, T. Birdal, H. Deng, and F. Tombari.
\newblock {3D point capsule networks}.
\newblock In \emph{CVPR}, 2019.

\bibitem[Zhou et~al.(2018)Zhou, Park, and Koltun]{zhou2018open3d}
Qian-Yi Zhou, Jaesik Park, and Vladlen Koltun.
\newblock Open3d: A modern library for 3d data processing.
\newblock \emph{arXiv preprint arXiv:1801.09847}, 2018.

\bibitem[Zhu et~al.(2023)Zhu, Zhang, He, Guo, Zeng, Qin, Zhang, and Gao]{Zhu_2023_ICCV}
Xiangyang Zhu, Renrui Zhang, Bowei He, Ziyu Guo, Ziyao Zeng, Zipeng Qin, Shanghang Zhang, and Peng Gao.
\newblock Pointclip v2: Prompting clip and gpt for powerful 3d open-world learning.
\newblock In \emph{ICCV}, pages 2639--2650, 2023.

\end{thebibliography}
}


\end{document}